\documentclass{article}

\usepackage{arxiv}

\usepackage[T1]{fontenc}    
\usepackage{hyperref}       
\usepackage{url}            
\usepackage{booktabs}       
\usepackage{amsfonts}       
\usepackage{nicefrac}       
\usepackage{microtype}      
\usepackage{lipsum}         
\usepackage{graphicx}
\usepackage{doi}
\usepackage{amsmath}
\usepackage{cleveref}       
\usepackage{enumitem}
\usepackage{multirow}
\usepackage{float}

\usepackage[style=authoryear, backend=biber, uniquename=false, uniquelist=false]{biblatex}
\setenumerate[2]{label=\alph*.}

\title{Fluent dreaming for language models}

\date{January 2024}

\author{ T. Ben Thompson, Zygimantas Straznickas, Michael Sklar \\
	Confirm Labs \\
	\texttt{t.ben.thompson@gmail.com}
}

\renewcommand{\headeright}{}
\renewcommand{\undertitle}{}
\renewcommand{\shorttitle}{Fluent Dreaming}

\hypersetup{
pdftitle={Fluent dreaming for language models},
pdfsubject={},
pdfauthor={T. Ben Thompson, Zygimantas Straznickas, Michael Sklar},
pdfkeywords={},
}

\begin{document}

\maketitle

\begin{abstract}
	Feature visualization, also known as "dreaming", offers insights into vision models by optimizing the inputs to maximize a neuron's activation or other internal component. However, dreaming has not been successfully applied to language models because the input space is discrete. We extend Greedy Coordinate Gradient, a method from the language model adversarial attack literature, to design the Evolutionary Prompt Optimization (EPO) algorithm. EPO optimizes the input prompt to simultaneously maximize the Pareto frontier between a chosen internal feature and prompt fluency, enabling fluent dreaming for language models. We demonstrate dreaming with neurons, output logits and arbitrary directions in activation space. We measure the fluency of the resulting prompts and compare language model dreaming with max-activating dataset examples. Critically, fluent dreaming allows automatically exploring the behavior of model internals in reaction to mildly out-of-distribution prompts. \footnote{Code for running EPO is available at \url{https://github.com/Confirm-Solutions/dreamy}. A companion page demonstrating code usage is at \url{https://confirmlabs.org/posts/dreamy.html}.}
\end{abstract}

\section{Introduction}

Feature visualization for vision models, also known as ``dreaming"\footnote{In this paper where we work exclusively with language models, we use "dreaming" to describe the task because "feature visualization" seems like an inappropriate term for a setting with no visual content.}, refers to a process of optimizing an input image to maximize the magnitude of a particular model feature. Feature visualization has been a powerful technique for vision model interpretability \parencite{cammarata2020thread, olah2018the, olah2017feature, mordvintsev-2015, yosinski2015understanding}. Feature visualization has not been successfully applied to text-only language models due to the difficulty of optimizing in the discrete space of text prompts. Attempts to run continuous optimization algorithms on relaxations using the Gumbel-Softmax approximation of the discrete sampling \parencite{jang2017categorical} have not been successful when applied to BERT \parencite{bauerle-2018} though moderate success has been seen on smaller models \parencite{poerner-etal-2018-interpretable}. 

In contrast, the automatic prompt engineering literature and the white-box adversarial attacks literature have both made steady progress towards optimizing prompts for language models \parencite{zou2023universal, wen2023hard, jones2023automatically, kumar2022gradientbased, shi2022human, shin2020autoprompt, ebrahimi2018hotflip}. In particular, the Greedy Coordinate Gradient (GCG) search method of \textcite{zou2023universal} is consistently successful at optimizing in token space to trigger objectionable content. In addition, progress has been made on "fluent" adversarial attacks where prompts with lower model-evaluated cross-entropy are preferred \parencite{zhu2023autodan, jones2023automatically, shi2022human}. As applied to language models, both ``white-box adversarial attacks'' and "dreaming" tasks optimize in the token space, use objectives which are typically differentiable with respect to input embeddings, and may optionally require linguistic fluency of the inputs. They primarily differ in the implied location of their targets - outputs in the case of``adversarial attacks'', and model internals in the case of ``dreaming''. 

Here, we modify and extend GCG to perform fluent dreaming for language models. First, we add a regularization term to the GCG objective function to prefer lower cross-entropy prompts. This encourages fluent prompts. However, the ideal regularization strength will depend on both the fluency desired by a human model interpreter and the context in which the feature activates. As a result, we lack a reliable a-priori rule for choosing a regularization strength. Instead, we extend GCG into an evolutionary algorithm in which each population slot optimizes for a different point on the Pareto frontier between cross-entropy and feature activation. We name the resulting algorithm Evolutionary Prompt Optimization (EPO). While we apply EPO to dreaming in this paper, EPO is likely useful in adversarial attacks and other token optimization settings where the goal is to optimize a prompt for both fluency and a differentiable objective.

Within text-only language models, efforts to interpret features commonly use max- or high-activation dataset examples \parencite{bricken2023monosemanticity, bills2023language}. While dreaming does overlap in purpose with max-activating dataset examples, we believe there are a few advantages of dreaming:

\begin{itemize}
    \item \textbf{Evaluation beyond the training set}:
    Dreaming can be useful when studying a feature where the most triggering prompts are not present in the training data \parencite{bolukbasi2021interpretability}. For example: Pythia 12B layer 10, neuron (see section \ref{sec:l10n5}) responds to "i.e." and "for example" but dreaming can find combinations such as "i.e. for example" which triggers the feature even more strongly. A dataset-based analysis fails to find such prompts. More broadly, real world or test data distributions often differ greatly from the training set. Sometimes train-test differences are adversarial in nature, especially in real world settings where users are interacting with an AI system.
    \item \textbf{Rare triggers}: Rare feature triggers are much more efficiently found by dreaming than by a max-activating examples analysis. We demonstrate this when we find that "exmaple" triggers Pythia 12B-L10.N5. We would need to scan a large dataset to find this trigger.
    \item \textbf{Higher activations}: We consistently achieve higher feature activations via dreaming than via max-activating dataset examples. 
    \item \textbf{Different algorithmic asymptotics}: The computational expense of dreaming is linear in the number of features investigated. In contrast, scanning a large dataset is a fixed cost which can be spread across thousands of features. Dreaming will be computationally much cheaper when we want to examine only a single feature or a small number of features.
\end{itemize}

US Executive Order 14110 of October 30, \parencite{executive14110} describes "red-teaming" for AI models as "[adopting] adversarial methods to identify flaws and vulnerabilities, such as harmful or discriminatory outputs from an AI system, unforeseen or undesirable system behaviors, limitations, or potential risks associated with the misuse of the system." The ability to produce fluent dreams may be advantageous for several of these tasks.

In the remainder of this paper, we introduce the EPO algorithm and demonstrate examples of optimizing the outputs and internals of Pythia-12B. 

\section{Evolutionary prompt optimization (EPO)}

Language model dreaming has several challenges that we solve:

\begin{itemize}
\item Past language model dreaming work has optimized in the model's continuous embedding space, but projecting this optimized embedding to the nearest token can produce low feature activation values \parencite{bauerle-2018}. While recent work in adversarial attacks may be solving this problem \parencite{yuan2023bridge}, optimization in embedding space remains challenging. Instead, we optimize in the discrete token space using an evolutionary algorithm based on GCG \parencite{zou2023universal} which uses token gradients to constrain the search space.
\item Prompts optimized solely to maximize a given feature will be gibberish because fluent language is a small manifold in prompt space and the optimizer has no constraint towards preferring fluent text. We use a language model to evaluate the fluency of the prompt by computing the self-cross-entropy of the prompt \parencite{jones2023automatically, zhu2023autodan, shi2022human}. This allows us to preference the search process towards reasonable text.
\item Methods from the adversarial attacks literature that regularize towards fluent text use a pre-specified regularization strength \cite{zhu2023autodan}, but the ideal regularization strength varies substantially depending on the feature that we are optimizing. Each population member in our evolutionary algorithm is chosen based on a different regularization strength. We construct a Pareto frontier of the tradeoff between prompt fluency and feature activation.
\end{itemize}

Given a feature $f(\mathbf{t})$ and a language model $m(\mathbf{t})$, we want to find a prompt $\mathbf{t}^*$ that maximizes the feature while minimizing the self-cross-entropy of the prompt. The total objective $L_{\lambda}(\mathbf{t})$ is:

\begin{align}
L_{\lambda}(\mathbf{t})&= f(\mathbf{t}) -  \frac{\lambda}{n} \sum_{i=0}^{n-1} H(m(\mathbf{t}_{\leq i}), t_{i+1}) \big) \\
\mathbf{t}^*_{\lambda} &= \underset{\mathbf{t}}{\mathrm{argmax}} ~ L_{\lambda}(\mathbf{t})
\end{align}

where $H$ is the cross-entropy operator, $\lambda$ is a weighting parameter determining the tradeoff between fluency regularizer and feature maximization, $n$ is the number of tokens in the prompt, $\mathbf{t}_{\leq i}$ is the prefix of $\mathbf{t}$ up to token $i$, and $t_i$ is the $i$-th token in $\mathbf{t}$.

Note that the feature under investigation, $f(\mathbf{t})$, must be differentiable with respect to a one-hot encoded token vector. Also $f$ may be a component of a different model from the model used for evaluating cross-entropy as long as the tokenization of the feature model and the language model are the same so that gradients can be propagated to the token space. We find that using larger models for evaluating cross-entropy, while more computationally expensive, produces dreams that are more understandable to a human interpreter.  

Because the ultimate goal is human interpretation of a feature, the ideal value of $\lambda$ can vary widely. Instead of choosing a single $\lambda$, our algorithm explores the Pareto frontier over a range of $\lambda$ values, allowing the user to select the desired tradeoff between fluency and feature maximization after running the algorithm. We optimize a family of objectives $L_{\lambda_i}$ parameterized by $M$ values of the fluency regularization strength, $\lambda_1, \dots, \lambda_M$.

To optimize this family of objectives, we use an evolutionary algorithm:

\begin{enumerate}
    \item Initialize (either user-provided or random) a population of $M$ prompts, $\mathbf{t}^{0},...,\mathbf{t}^{M}$ each of length $n$. The prompt lengths are fixed for the duration of the algorithm, however there is nothing about the algorithm that inherently requires constant-length prompts.
    \item Compute the feature magnitude and cross-entropy of each prompt. Then following the approaches of \parencite{ebrahimi2018hotflip, shin2020autoprompt, zou2023universal}:
    \begin{enumerate}
        \item For each member of the population, backpropagate to a one-hot encoding of the prompt to obtain a gradient for each token in the vocabulary in each token position: $\nabla_{e_{x_i}} L_{\lambda}(\mathbf{t})$.
        \item Select the top-$k$ tokens in each token position according to the magnitude of the gradients computed in the prior step.
        \item For each member of the population, generate $r$ children by replacing a single token position with one of the top-$k$ tokens in that position. The token position is chosen uniformly at random and the new token is chosen uniformly at random amongst the top-$k$ tokens.
    \end{enumerate}
    \item We now have $Mr$ prompts. Select the best prompt with replacement according to each of the $L_{\lambda_i}$.\footnote{Selecting the best prompt \textit{without replacement} performed substantially worse.} The result is a population of $M$ prompts. 
\end{enumerate}

The basic EPO algorithm repeats step 2 and 3 for $T$ iterations. Note that GCG \parencite{zou2023universal} is equivalent to EPO with $M=1$ and $\lambda=0$.

We find that EPO and GCG frequently get stuck in local minima. To enhance exploration, we semi-randomly "restart" the population every $T_{\mathrm{restart}}$ iterations. We choose the hyperparameters $\lambda_{\mathrm{r,min}}$ and $\lambda_{\mathrm{r,max}}$. To perform a restart, we select a $\lambda_{r}$ uniformly at random in $[\lambda_{\mathrm{r,min}}, \lambda_{\mathrm{r,max}}]$. Then, we select the optimal population member according to $L_{\lambda_{r}}$ and remove all the other population members. Immediately after a restart, the Pareto frontier will be substantially worse because it will consist of only the single retained example. However, by 10 iterations after a restart, solutions are typically substantially better at all points on the Pareto frontier.

Compared to existing fluent prompt optimization algorithms like \textcite{jones2023automatically, zhu2023autodan}, the addition of a population of multiple candidate prompts and the use of a family of fluency regularization strengths allows us to explore the full Pareto frontier of fluency and feature maximization. The presence of multiple regularization strengths and restarting also helps to prevent the algorithm from getting stuck in local optima. 

In the left-to-right AutoDAN adversarial attack algorithm, the fluency regularization term is computed separately from the gradient in the token selection step 2a. While this approach probably will probably make the fluency regularization more reliable, we prefer to avoid the additional hyperparameter that results. Instead, we compute the gradient of the full objective including the fluency term.

The remainder of this paper will focus on demonstrating that EPO can achieve successful, fluent dreaming. We leave to future work an in-depth study of EPO's algorithmic design choices and comparisons to other methods.

\section{Dreaming experiments}

We apply EPO-based dreaming to various optimization objectives defined on the internals and outputs of Pythia-12B \parencite{biderman2023pythia}. For simplicity, we also use Pythia-12B for evaluating fluency.\footnote{We chose Pythia-12B for historical reasons. If we were beginning the project now, we would consider smaller and more capable LLMs for ease of use reasons.}

We begin in Subsections \ref{subsec:logits} and \ref{subsec:neurons} with a broad exploration of dreaming outputs, investigating maximal token logits and neuron activations. In Subsection \ref{subsec:residual}, we explore the quantitative performance of EPO in optimizing residual stream activations.  

The EPO hyperparameters we use are shown in Table \ref{tab:parameters}. We exclusively use 12-token-long randomly initialized prompts. One important note is that the mean cross-entropy of a 12-token slice of the Pile is 3.7 and the standard deviation is 1.1 (Figure \ref{fig:residual_alignment}a). Thus, two standard deviations above mean cross-entropy is 5.9 and prompts with a cross-entropy below ~6 can reasonably be considered to be within-distribution. The $M=8$ values of $\log(\lambda_i)$ are uniformly gridded between $\log(1/10)$ and $\log(10)$. 

\begin{table}[h]
\centering
\begin{tabular}{|c|c|}
\hline
Parameter & Value \\
\hline
$T$       & 300   \\
$M$       & 8    \\
$r$       & 32     \\
$k$       & 512    \\
$\lambda_{r,min}$ & 0.667 \\
$\lambda_{r,max}$ & 6.0 \\ 
$T_{\mathrm{restart}}$ & 30 \\ 
\hline
\end{tabular}
\caption{Parameter Values}
\label{tab:parameters}
\end{table}

\subsection{Token logits}
\label{subsec:logits}

Maximizing a class logit or probability has been a common objective function for computer vision dreaming work \parencite{olah2017feature}. We aim to find prompts that maximize the output probability of a particular token. This objective is an easier form of an adversarial attack objective that forces the model to output a particular sequence. Nonetheless, maximizing class logits provides a good test bed for demonstrating fluency. The precise objective we choose to optimize for is the difference in logits between the target token and the most likely alternative token:
\begin{equation}
    m(\mathbf{t})_{g} - \underset{i \neq g}{\mathrm{argmax}} ~ m(\mathbf{t})_i
\end{equation}
where $g$ is the target (goal) token.

\begin{table}
\caption{Pareto frontier examples from optimizing to maximize token logits. A positive logit difference implies that the target token is the most likely next token according to Pythia-12B.}
\label{tab:logit}
\begin{tabular}{lrrl}
\toprule
 & Logit Difference & Perplexity & Text \\
Target token &  &  &  \\
\midrule
dog & -1.58 & 72.05 &  animaux can range from those that can quickly startle a \\
    & 0.97 & 91.08 &  animaux can range from those that can quickly startle your \\
    & 1.13 & 93.97 &  animaux can range from those that can quickly rile your \\
    & 3.24 & 207.67 &  canine not running is shown by 1. >The \\
    & 3.59 & 341.05 &  canine not running and shown by 1. >The \\
    & 4.57 & 6,717.69 &  canine not running="\{\{ shown standing1. > The \\
    & 4.62 & 18,840.23 &  canine reading Bh triv (`/ de\textbackslash > The \\
\hline
AI & -6.66 & 45.53 &  SIIS-A/B as opposed to the one that \\
   & -1.22 & 67.95 &  SIIS-A/B as opposed to the what the \\
   & 2.85 & 128.44 &  understand Mori's code. It's data, everything an \\
   & 3.09 & 174.87 &  understand Mori's code. That's data, everything an \\
   & 3.77 & 1,058.75 &  weaving Chloe Glor's code. It's data; everything an \\
   & 3.98 & 1,858.16 &  SIOL-A/* ABA Contrary to other Players the \\
   & 4.54 & 10,989.41 &  AKIIS \{\{ sendStateAO \} Conversely to real players the \\
\hline
grand & -2.25 & 46.79 &  Community Center and Northwinds ISD, speaks during the \\
      & -2.22 & 46.88 &  Community Center and Northwinds ISD, speaks during the \\
      & -0.88 & 56.77 &  Community Center and Northwinds AFC, speaks during the \\
      & 3.48 & 115.13 &  Retail Center, Tradewinds Kona, speaks during its \\
      & 4.23 & 210.94 &  Retail Center retailer Tradewinds Kona, speaks during its \\
      & 4.50 & 411.39 & Building Center retailer Tradewinds Kona, speaks during its \\
      & 4.87 & 1,481.46 &  Care Store of Broad Fair Holdings Kona, speaks during its \\
\bottomrule
\end{tabular}
\end{table}

In addition, during the optimization, we reject any token in the prompt that is a variation on the target token. So, if the target token is "dog", we reject any prompts that include the character sequence "dog", upper or lowercase. Without this constraint, the optimization often just repeats the target token several times in the prompt.

We run EPO twice for each target token. We share the Pareto frontier results from optimizing for the output tokens "AI", "dog" and "grand" in Table \ref{tab:logit}. We achieve fluent dreams as determined both by perplexity calculations using Pythia-12B and visual inspection. While the perplexity of these examples may seem high in comparison to typical language model loss, it's important to remember that the prompts here are only 12 tokens and a perplexity of 121 is one standard deviation above the mean in the training distribution. As expected, there is an inverse relationship between logit difference and fluency (perplexity). 

\subsection{Multilayer perceptron neurons}
\label{subsec:neurons}

We apply EPO to a range of multilayer perceptron (MLP) neurons in Pythia-12B. While neurons are likely to be polysemantic \parencite{szegedy2014intriguing, elhage2022superposition}, there may be a small number of monosemantic neurons or neurons which are monosemantic above a certain activation threshold \parencite{gurnee2023finding}. We refer to neuron X in layer Y as "LY.NX". For example, "L5.N1035". For an unbiased perspective on the outputs, we run EPO once on each of six neurons. We share Table \ref{tab:six-examples} containing the maximum activating prompt with perplexity below 100 for each neuron. We also share all the Pareto-optimal prompts in an appendix in Table \ref{tab:six-examples-appendix}.

\begin{table}[h]
\caption{Results of running EPO on six neurons. The maximum activating prompts produced by EPO with perplexity below 100.}
\label{tab:six-examples}
\begin{tabular}{llrrl}
\toprule
 &  & Activation & Perplexity & Text \\
layer & neuron &  &  &  \\
\midrule
2 & 0 & 2.59 & 81.64 & HRC) and other discoursing agencies. In Chap \\
7 & 1 & 0.22 & 41.37 & ”). The reason why Dadaab was created, and the \\
12 & 2 & 4.95 & 84.56 &  HELSTON... but has not escaped entirely un \\
17 & 3 & 3.03 & 93.60 &  Theater® is the interactive hit on TV, streaming online or \\
22 & 4 & 2.60 & 85.56 & BURG Code and SA 2000-01:9, a \\
27 & 5 & 3.02 & 61.87 &  wont get it. I think Jace is the surprise m \\
\bottomrule
\end{tabular}
\end{table}

\begin{figure}[h]
  \centering
  \includegraphics[width=0.5\linewidth]{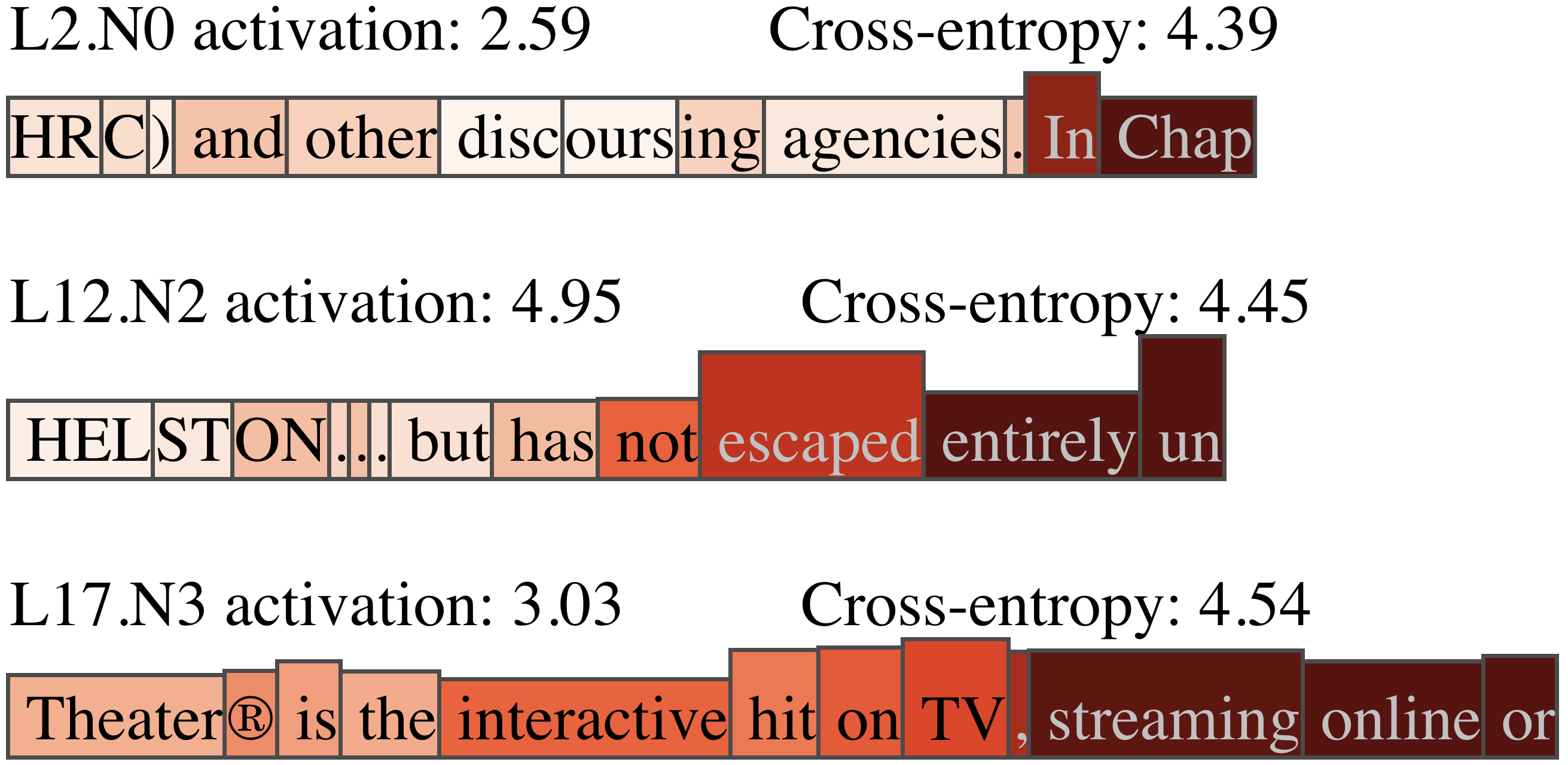}
  \caption{Token-level attribution for three of the examples from Table \ref{tab:six-examples}. For each token position, we use the same top-k gradient operation used in GCG and EPO to identify 32 candidate replacement tokens. The color of the token in the visualization corresponds to the reduction in activation from the worst substituted token in that position. Dark reds indicate that a different token in that position can reduce the activation to almost zero. The height of the token bar indicates the reduction in activation from the best alternative token in that position. Tall bars indicate that all potential token substitutions reduce activation dramatically and thus the precise token is very important. We share interactive versions of this visualization at \url{https://confirmlabs.org/posts/dreamy.html}.}
  \label{fig:attribution}
\end{figure}

We also share causal token attribution for three of the examples in Figure \ref{fig:attribution}. The layer 2 neuron responds to tokens near the end of the prompt, while the layer 12 neuron and especially the layer 17 neuron respond to earlier context.

\begin{figure}[h]
  \centering
  \includegraphics{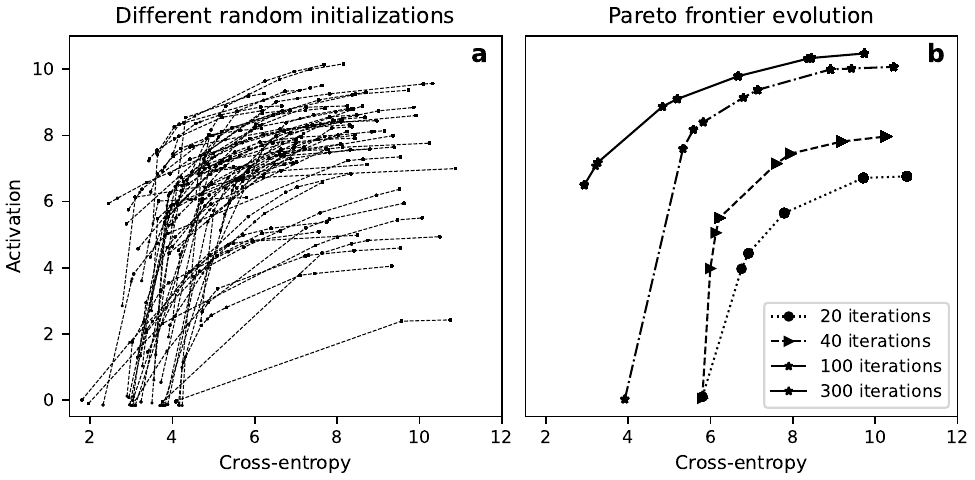}
  \caption{\textbf{a}) Pareto frontiers for sixty runs of EPO applied to L10.N5 with different random initializations. \textbf{b}) The evolution of the Pareto frontier during a single run of EPO.}
  \label{fig:evolution_and_random}
\end{figure}

In Figure \ref{fig:evolution_and_random}a, we show the Pareto frontier from 60 runs of EPO applied to L10.N5 and randomly initialized with different seeds. There is large variation in the final state of the Pareto frontier across different runs, with maximal activation varying from 2.4 to 10.2 and maximal activation with cross-entropy less than 5 varying from 0 to 8.6.

In Figure \ref{fig:evolution_and_random}b, we show the evolution of the Pareto frontier at different time points during the EPO algorithm. The Pareto frontier is progressively pushed outwards. Much of the progress in the later iterations occurs at lower cross-entropy levels. 

\subsection{Pythia-12B-L10.N5, the "example" neuron}
\label{sec:l10n5}

In this section, we explore a particular neuron, L10.N5, using the results from 60 runs of EPO. L10.N5 reacts strongly to a variety of phrases related to examples. Activations greater than 5 seem to be almost always triggered by example-related phrases. See Table \ref{tab:examples}. We discuss a few types of prompts that activate L10.N5:

\begin{table}
\caption{Examples of EPO-generated prompts that strongly activate L10.N5.}
\label{tab:examples}
\begin{tabular}{lrrl}
\toprule
Activation & Cross-Entropy & Text \\
\midrule
\textbf{Basic phrases} \\
\midrule
7.34 & 3.48 &  Use this to program to a specific frequency, so for example \\
6.29 & 3.49 & Cookies are for your domain or path. ex. if \\
5.93 & 4.95 &  immunohistochemical staining performed at the specified primary \\
& & focus in parentheses (eg \\
5.37 & 4.95 &  primers-for primer of interest-specific.(i.e \\
6.52 & 4.95 &  Jenni Maree, use the percent (so if \\
7.64 & 9.66 &  active serum \textbackslash *H or Enlishley reported alterations (* ejemplo \\
\midrule
\textbf{Adversarial} \\
\midrule
6.71 & 3.81 &  priced in relation to the previous price (ie example, if \\
8.45 & 6.96 &  appellees were
each individually operated - example fo
ple, \\
8.85 & 7.25 &  arrows kindly headed link to a topic, for EX'd example \\
7.43 & 8.02 &  Created
// any percentage omissions, example f.ie. \\
\midrule
\textbf{Misspelling} \\
\midrule
5.04 & 3.89 & company-phone-number,email-address
Exsample \\
8.06 & 6.85 & Select Correspond with  Source    ( For    exampl, \\
\midrule
\textbf{Max-activation} \\
\midrule
10.16 & 8.16 &  following column:
Thedate*KKril\%). example, \\
10.13 & 7.68 & Longlowball in percent apart used number (so for example \\
\bottomrule
\end{tabular}
\end{table}

\begin{itemize}
\item \textbf{Simple example-related phrases}: The neuron reacts to common example-related phrases: "example", "for example", "i.e"., "eg", "so if". The neuron is also bilingual and reacts to "ejemplo". 
\item \textbf{Misspellings}: We see high activations for prompts with mild misspellings like "Exsample" and "For exampl,". The neuron also responds strongly to "exmaple" and "For exam\textbackslash n ple".
\item \textbf{Adversarial attacks}: strong reactions to "ie example, if" and "for EX'd example" suggest that the neuron is checking for the presence of several phrases and included multiple such phrases increases neuron activation. Other examples of the same type: "ie for example", "so for eged example", "for eg", "ie for example", "example f.ie.", "example fo\textbackslash nple". Some of these prompts would be very unlikely to be found using a dataset-driven approach.
\end{itemize}

While many of these insights into L10.N5 would be possible with dataset examples, optimization provides us with adversarial prompts that would be almost impossible to find in the training data. These prompts can inform mechanistic interpretation of the feature. The prompt containing "example f.ie." is most likely an adversarial attack on circuitry that responds to "example" and "i.e."

Despite the interesting behavior, the analysis above neglects the polysemanticity of L10.N5. There are many prompts that cause L10.N5 to produce activations up to ~5. It is only for activation values above 5 where the neuron is primarily reacting to example-related phrases. 

\begin{figure}[h]
  \centering
  \includegraphics[width=0.65\linewidth]{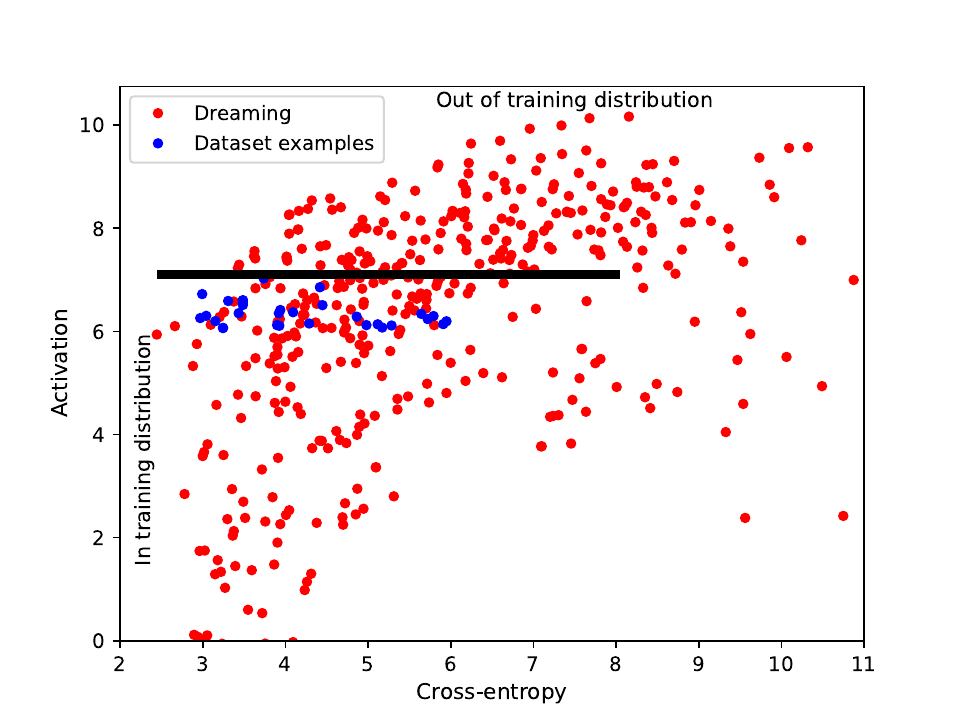}
  \caption{Comparing activation and cross-entropy between dreaming outputs and the top 64 max-activating dataset examples from 500 million tokens of the Pile. The black line is schematically separating regions of the plot that are empirically inside and outside the training distribution.}
  \label{fig:l10n5_dreaming_vs_dataset}
\end{figure}

In Figure \ref{fig:l10n5_dreaming_vs_dataset}, we compare the activations and cross-entropies between dataset examples and dreaming. The dreaming distribution covers the in-distribution dataset example region, but also extends far outside to higher activations and higher cross-entropies. At a cross-entropy of ~4, dreaming is producing activations 1-1.5 units higher than found in the training distribution.

\subsection{Pythia-12B-L3.N1000}

We show the results of applying dreaming to L3.N1000 (neuron 1000 in layer 3). This neuron focuses very heavily on the last few tokens in the prompt, a common feature of early layer neurons. In contrast to L10.N5, we see polysemanticity even in the highest activation values. The behavior is not easily decipherable.

\begin{enumerate}
    \item The neuron may respond to component-related words like "nodal", "modules", "element" and "subsystem".
    \item The neuron responds strongly to list items like "1:", "2:" and so on. 
    \item The word "depending" is causing a strong reaction. 
\end{enumerate}

\begin{figure}[h]
  \centering
  \includegraphics[width=0.65\linewidth]{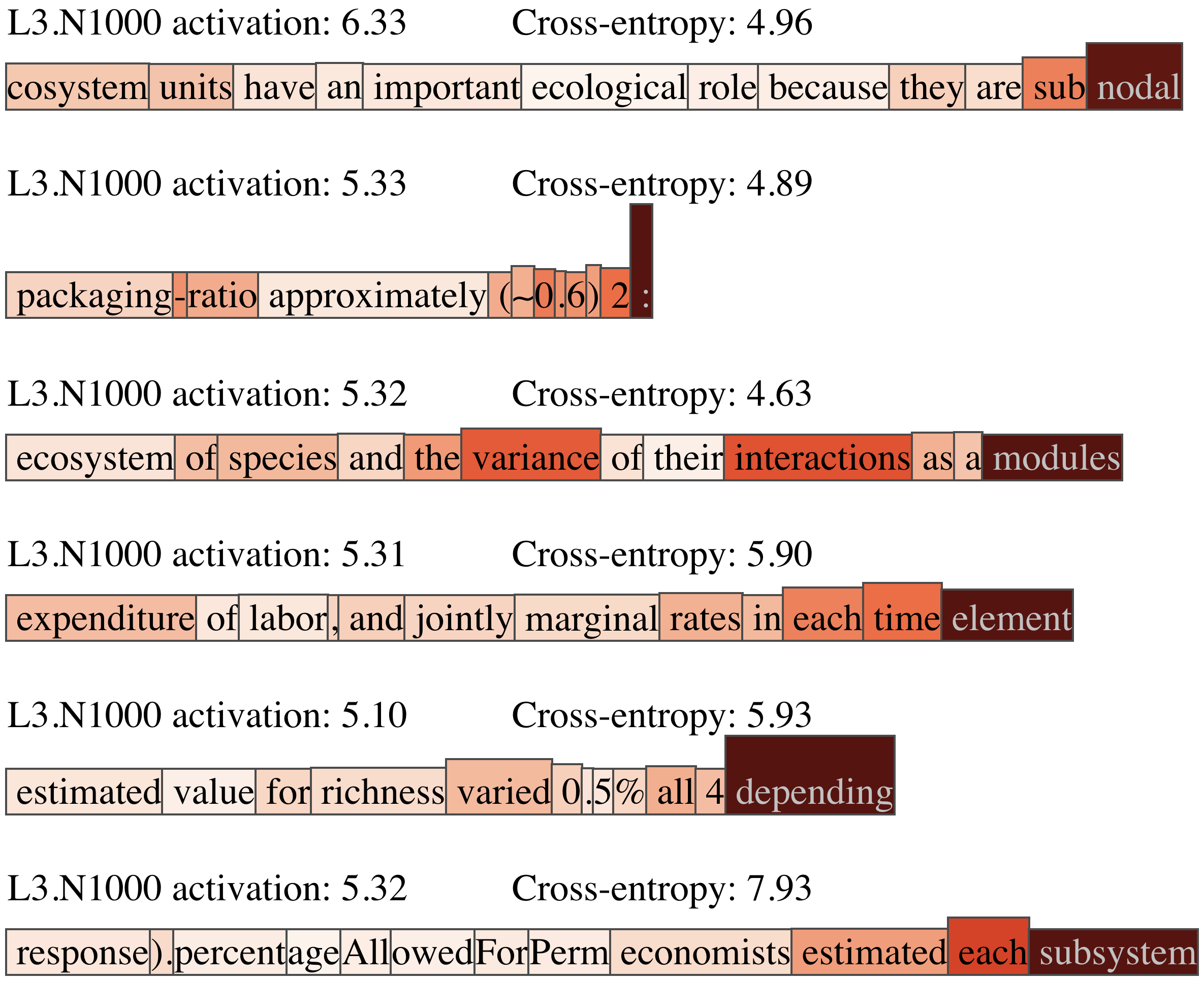}
  \caption{Causal attribution visualizations for six prompts produced by applying dreaming to L3.N1000. See the caption of Figure \ref{fig:attribution} for an explanation of the attribution method.}
  \label{fig:attribution_L3N1000}
\end{figure}

While applying dreaming to this neuron only provides a small amount of evidence regarding its function, the activations achieved by dreaming are higher than those found via dataset examples. The max-activating dataset example has activation 4.6 and cross-entropy 4.3. 

\subsection{Residual Stream}
\label{subsec:residual}

In this section, we explore the degree to which we can affect random directions within the activation space of a language model. Specifically, we study the residual stream \parencite{elhage2021mathematical} at different layers. The residual stream dimension of Pythia-12B is 5120. We choose a random vector in this space, $v$, and maximize the alignment of the normalized residual stream with that vector:

\begin{equation}
\label{eqn:alignment}
    \bigg(\frac{\mathbf{x}_L(\mathbf{t}) - \mu_{\mathbf{x}_L}}{\sigma_{\mathbf{x}_L}}\bigg)^Tv 
\end{equation}

where $\mathbf{x}_L(t)$ are the residual stream activations at layer $L$ given the input prompt $t$. 

We compare three approaches to optimizing this objective:
\begin{enumerate}
    \item Sample 1024 prompts with each token chosen uniformly at random.
    \item Scan 100 million random tokens of the Pile and find the prompts that maximize the above objective.
    \item Apply EPO.
\end{enumerate} 

We select 50 random vectors in activation space at each of seven layers (4, 8, 12, 16, 20, 24, 28). Then, we apply the three approaches above to maximize the alignment with the random vector according to Equation \ref{eqn:alignment}. It is much easier to align the residual stream with some random vectors than others. In order to remove that variation, we use the results from method 1, the random prompt approach, to standardize the alignment scores. We compute the mean alignment, $\mu^{\mathrm{random}}_{i, L}$, and the standard deviation, $\sigma^{\mathrm{random}}_{i, L}$ across the 1024 prompts for each of the 50 vectors in each layer. For the rest of this section, we abuse notation and leave the $i$ and $L$ implicit. We use these values to construct a z-score metric that measures how many standard deviations from the mean an activation would be in the distribution of activations resulting from random prompts:

\begin{equation} 
z = \frac{a - \mu_{\mathrm{random}}}{\sigma_{\mathrm{random}}}
\end{equation}

where $a$ is the alignment.

\begin{figure}[h]
  \centering
  \includegraphics[width=\linewidth]{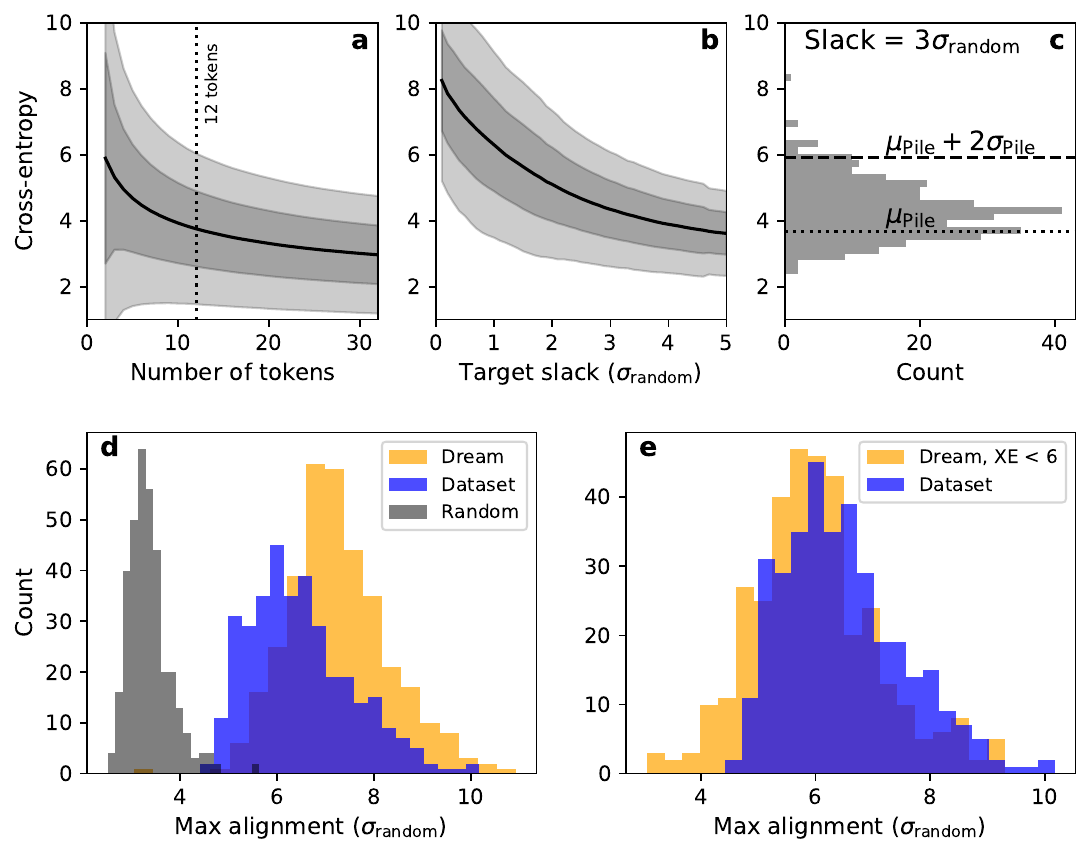}
  \caption{\textbf{a}) Average cross-entropy plotted against number of tokens for text in the Pile. The dark gray region shows one standard deviation above and below average cross-entropy while the light gray region shows two standard deviations. Note that all the optimized prompts in this paper are 12 tokens in length. \textbf{b}) We plot average cross-entropy across the residual vector alignment dreaming runs as a function of slack below maximum alignment as measured in units of random prompt standard deviations. See the text for a discussion of the units. \textbf{c}) The distribution of cross-entropy for dreaming results with a slack of three standard deviations.\textbf{d)} The distribution over 50 random target vectors of maximum alignment for each method. \textbf{e}) Similar to d but, in order to approximately restrict the analysis to in-distribution prompts, we restrict the dreaming outputs to have cross-entropy below 6. Note that we do not restrict the cross-entropy of the max-activating dataset prompts because they are known to be in-distribution.}
  \label{fig:residual_alignment}
\end{figure}

In Figure \ref{fig:residual_alignment}d, we plot the distributions of these maximum alignment z-scores over the 50 random target vectors between our three different methods. We can see that dreaming results in the highest activations when we have no cross-entropy constraints. This matches the results in \ref{fig:l10n5_dreaming_vs_dataset} where we see that dreaming is able to sample out-of-distribution prompts. The random prompts are unable to reach very high alignments. In Figure \ref{fig:l10n5_dreaming_vs_dataset}e, we plot the distribution of maximum alignment restricting the dreaming outputs to have cross-entropy less than 6. In this case, the alignments found through dataset scanning tend to be slightly higher than through dreaming, but the distributions are broadly similar. 

We would like to establish the fluency of dreaming outputs. In Figure \ref{fig:residual_alignment}b, we plot the minimum cross-entropy of the prompts as a function of the slack below the overall maximally aligned prompt. For example, at a slack of 2.0, we select all the prompts with alignment $2\sigma_{random}$ below the maximally alignment and then find the minimum cross-entropy amongst those prompts. This figure represents an averaged view of the Pareto frontier tradeoff between cross-entropy and activation. In Figure \ref{fig:residual_alignment}c, we can see that, for a slack of $3\sigma_{random}$, the distribution of dreaming prompts is very similar to the Pile.

\begin{figure}[h]
  \centering
  \includegraphics[width=0.4\linewidth]{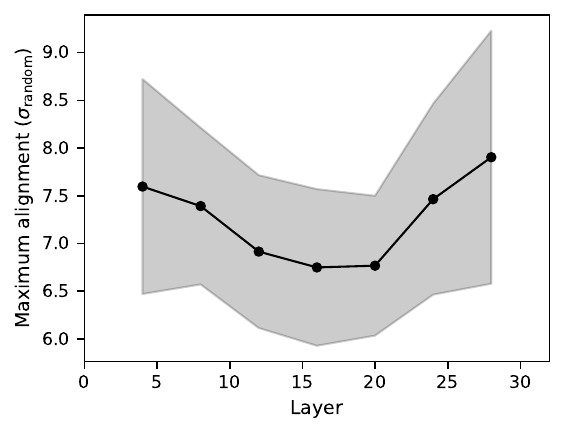}
  \caption{We apply EPO to optimize residual stream alignment for 50 random vectors in each of the plotted layers. We then compute the mean and standard deviation of the maximum alignment z-score.}  
  \label{fig:layer_alignment}
\end{figure}

In Figure \ref{fig:layer_alignment}, we plot the mean and standard deviation of maximal alignment from EPO across the layers of Pythia-12B in units of $\sigma_{\mathrm{random}}$. This shows that EPO is consistently able to influence the residual stream across all the layers of the model. The alignment is substantially lower in the middle layers of the model. Unsurprisngly, EPO is many orders of magnitude more efficient than random sampling. Drawing a sample that lies 7.5 standard deviations from the mean of a Gaussian distribution would require, in expectation, around one trillion draws.

\section{Polysemanticity}

Both language models neurons and non-basis-aligned vectors in activation space are frequently polysemantic \parencite{gurnee2023finding, elhage2022superposition, szegedy2014intriguing}. That is, a neuron or vector may activate for multiple distinct and often anti-correlated features, concepts or prompts. Recent work has made progress on disentangling polysemanticity within language models, constructing a set of mostly monosemantic features by bottle-necking internal model states through a sparse autoencoder \parencite{cunningham2023sparse, bricken2023monosemanticity}. While we have chosen not to apply dreaming to the features output by dictionary learning, we believe that would be a very valuable direction to extend this work.

Dreaming as a tool for intepretability can be misleading when the highest-activating prompts are not representative of typical functioning \parencite{bolukbasi2021interpretability}. Or, discrete search may simply happen to miss important parts of the prompt space. Attempts to explain neurons often take various percentile-activating dataset examples into account, in addition to max-activating dataset examples \parencite{olah2018the}.

Optimizing for diversity has been successfully applied in vision model feature visualization \parencite{olah2017feature} and similar techniques might be useful to identify less-than-maximally activating prompts for language model features. There is a mild natural tendency towards diversity in our current algorithm because the model gets stuck in very different local minima for different random seeds. We would be particularly excited about extensions of our algorithm that encourage mechanistic diversity. That is, an algorithm that finds a range of prompts that activate the same feature via different upstream internal model circuitry. Methods based on measuring composition between neurons and attention heads \parencite{elhage2021mathematical} could be used to measure mechanistic diversity.


\section{Hyperparameters}

Hyperparameter tuning is very important for performance of EPO. With poor hyperparameter, EPO fails to achieve fluency. However, we have not performed systematic hyperparameter searches with EPO. The parameters we use here are roughly designed to favor variety and exploration instead of fast convergence. Applying GCG or EPO in other settings may benefit from very different hyperparameters. For example, in our work on adversarial attacks, we use $k=32$ instead of $k=512$. In our rough experience, using a lower value of $k$ results in faster convergence but significantly less fluent prompts. An optimal implementation would likely increase $k$ as the number of iterations increases. Early in the algorithm, small $k$ results in fast convergence towards a basin of attraction. Late in the algorithm, large $k$ allows enough exploration to continue improving the solutions. The hyperparameters we use here are designed for a target on the same scale as the cross-entropy. Optimization targets on different scales will require different hyperparameters or scaling. 

EPO is computationally slow because it requires a forward pass of the model for each candidate prompt. With the hyperparameters we have used here, a single optimization job runs in about 200 seconds on a single RTX A6000 GPU. In general, discrete optimization is quite expensive, but we are optimistic about future algorithmic improvements for dreaming.

\section{Conclusions}

In this paper, we presented the first successful algorithm for fluent dreaming in language models. Our algorithm, Evolutionary Prompt Optimization (EPO) supplements discrete search methods with token gradients to construct a Pareto frontier between fluency and feature activation. We demonstrate EPO producing high-activation prompts that maintain linguistic coherence. Dreaming with EPO is especially valuable as a technique for exploring out-of-distribution behavior inside a language model.

We see a variety of potential improvements and extensions to EPO. EPO and other algorithms based on single token swaps get trapped in local minima when optimizing for fluency. This is especially true when an important single word is composed of two tokens. Swapping either token alone will dramatically reduce fluency and may not have a countervailing increase in activation. Swapping both tokens, on the other hand, could improve fluency and activation. Modifications to EPO that allow insertion and deletion of tokens could be another fruitful direction. Like GCG \parencite{zou2023universal}, our algorithm only evaluates the gradient at the current sequence. However, ARCA \parencite{jones2023automatically} suggests the possibility of averaging gradients at multiple neighbors to improve the quality of first-order approximation. In general, GCG-based methods can get stuck in local minima very quickly. (see Figure \ref{fig:evolution_and_random}a). We are also interested in algorithmic modifications that would result in more exploration. Methods that increase exploration would hopefully result in more consistent optimization results when initialized from different random seeds. 

EPO is likely to be valuable well beyond fluent dreaming.
The prompt fluency benefits of EPO may aid adversarial attacks for red-teaming by evading perplexity-based filters or other basic oversight techniques. EPO can also support automatic prompt engineering or optimization of any function of the prompt that is differentiable with respect to the embedding. Algorithms similar to EPO may also be useful for controllable generation similar to \textcite{kumar2022gradientbased}. In sum, EPO is a versatile choice for fluently optimizing a language model's inputs to maximize a differentiable objective. 

\section*{Acknowledgements}

We thank Martin Wattenberg for an early conversation that inspired this work.

\newpage
\printbibliography

\newpage
\appendix
\section{Additional Tables}

\begin{table}[H]
\caption{Results of running EPO on six neurons.}
\label{tab:six-examples-appendix}
\begin{tabular}{llrrl}
\toprule
 &  & Activation & Perplexity & Text \\
layer & neuron &  &  &  \\
\midrule
\multirow[t]{8}{*}{2} & 0 & -0.05 & 42.36 & HRC) and other discoursing members held a meeting \\
 & 0 & 2.54 & 79.13 & HRC) and other discoursing organizations. In Chap \\
 & 0 & 2.59 & 81.64 & HRC) and other discoursing agencies. In Chap \\
 & 0 & 3.93 & 262.52 & HRC) and other discoursing agencies". In \\
 & 0 & 4.71 & 549.27 & HRC) and other discoursing organizationsPosted In \\
 & 0 & 5.89 & 2,056.80 &  Record Company) and other discoursing folkspackage.[\textasciicircum  \\
 & 0 & 6.22 & 9,698.12 &  Record Company) recwith discoursing pleapackage][\textasciicircum  \\
 & 0 & 6.71 & 156,519.75 & OPLE COURT) st AdditionalAttoursshinelicense CopyrightSent \\
\cline{1-5}
\multirow[t]{7}{*}{7} & 1 & 0.14 & 40.18 & ”. The reason why Dadaab was created, and the \\
 & 1 & 0.22 & 41.37 & ”). The reason why Dadaab was created, and the \\
 & 1 & 1.98 & 111.59 &  ” The reason why Dadaab was created initially — the \\
 & 1 & 2.45 & 187.61 & …” The mystery how a thought actually was created, on what \\
 & 1 & 2.54 & 232.58 &  ” The reason why Dadaab was created 1919 — the \\
 & 1 & 3.13 & 1,030.19 & .** TheReason behind Dada\_ was established originally and registered \\
 & 1 & 3.41 & 3,099.71 & ''' The reason why Horus actually was awaited‐’ what \\
\cline{1-5}
\multirow[t]{7}{*}{12} & 2 & 4.61 & 57.22 & KESTON.... has not escaped entirely un \\
 & 2 & 4.95 & 84.56 &  HELSTON... but has not escaped entirely un \\
 & 2 & 5.11 & 106.07 &  MORESTON... but has not escaped entirely un \\
 & 2 & 6.00 & 642.16 &  TeV. Yet our own modest sum was not entirely escaped un \\
 & 2 & 6.32 & 1,297.21 &  TeV. Yet our own modest on was not entirely escaped un \\
 & 2 & 6.75 & 5,928.34 &  WESTINGTON ration per-It hardware not escaped entirely un \\
 & 2 & 6.78 & 14,558.60 &  WEabINGTON ration per-It hardware not escaped entirely un \\
\cline{1-5}
\multirow[t]{12}{*}{17} & 3 & 0.13 & 30.15 &  STV is the best format on TV, the radio and \\
 & 3 & 2.10 & 48.66 & OPLE TODAY is the newest hit on TV, streaming online and \\
 & 3 & 2.55 & 55.89 & Quest™ is the newest hit on TV, streaming online or \\
 & 3 & 2.60 & 57.22 &  Books™ is the newest hit on TV, streaming online or \\
 & 3 & 2.84 & 65.86 &  Theatre™ is the newest hit on TV, streaming online or \\
 & 3 & 2.90 & 73.18 &  Theater® is the latest hit on TV, streaming online or \\
 & 3 & 3.00 & 87.93 &  Underground™ is the interactive hit on TV, streaming online or \\
 & 3 & 3.03 & 93.60 &  Theater® is the interactive hit on TV, streaming online or \\
 & 3 & 3.18 & 147.83 &  Museum™ is the interactive hit on TV, streaming online or \\
 & 3 & 3.54 & 3,595.72 &  UCLA section, The CW incarnate ABC form currently 24 USA \\
 & 3 & 3.61 & 6,770.38 &  UCLA section, The CW incarnative ABC form currently 24 USA \\
 & 3 & 3.62 & 7,853.78 &  UCLA work, The CW incarnative ABC form currently 24 USA \\
\cline{1-5}
\multirow[t]{4}{*}{22} & 4 & 2.60 & 85.56 & BURG Code and SA 2000-01:9, a \\
 & 4 & 4.84 & 377.51 &  Wendel Code and SA 24-9:9 limited a \\
 & 4 & 5.18 & 542.87 & Keel Code and SA 2000-7:9 limited a \\
 & 4 & 5.41 & 1,665.65 &  Iz VII Code and SA 2000-179:14 limited a \\
\cline{1-5}
\multirow[t]{7}{*}{27} & 5 & -0.11 & 24.80 &  wont get it. I think Jace is the best. \\
 & 5 & 2.27 & 28.94 &  wont get it. I think Jace is the best m \\
 & 5 & 2.66 & 38.12 &  wont win it. I think Jace is the best m \\
 & 5 & 3.02 & 61.87 &  wont get it. I think Jace is the surprise m \\
 & 5 & 3.30 & 100.81 &  wont get taken. I think Jace is the surprise m \\
 & 5 & 4.52 & 1,204.42 &  naive poker is. ALL predict JLO being the first m \\
 & 5 & 4.59 & 1,558.63 &  selection poker is. ALL predict JLO being the first m \\
\cline{1-5}
\bottomrule
\end{tabular}
\end{table}

\end{document}